# *CLEAN*ANERCorp: Identifying and Correcting Incorrect Labels in the ANERcorp Dataset

**Mashael Al-Duwais, Hend Al-Khalifa and Abdulmalik Al-Salman**
King Saud University
Riyadh, Saudi Arabia
{malduwais, hendk, salman}@ksu.edu.sa

**Abstract**
Label errors are a common issue in machine learning datasets, particularly for tasks such as Named Entity Recognition. Such label errors might hurt model training, affect evaluation results, and lead to an inaccurate assessment of model performance. In this study, we dived deep into one of the widely adopted Arabic NER benchmark datasets (ANERcorp) and found a significant number of annotation errors, missing labels, and inconsistencies. Therefore, in this study, we conducted empirical research to understand these errors, correct them and propose a cleaner version of the dataset named CLEANANERCorp. CLEANANERCorp will serve the research community as a more accurate and consistent benchmark.

## 1. Introduction

Named Entity Recognition (NER) is the task of identifying both spans and types of named entities in text. It is a fundamental task in the natural language processing pipeline.

The ANERcorp dataset is the most well-known and utilized dataset for Arabic NER (Benajiba et al., 2007) and is a crucial benchmark for evaluating Arabic NER approaches. ANERcorp consists of 316 manually annotated articles from the news domain.

Deep Learning approaches have achieved state-of-the-art performance in the ANERcorp dataset with F1-score (0.84, 0.88, 0.89, 0.91, 0.92) (Al-Qurishi & Souissi, 2021; Alsaaran & Alrabiah, 2021; Antoun et al., 2021a, 2021b; Khalifa & Shaalan, 2019) respectively.

However, these experiments did not consider all tags during their experiments and used different data splits. This poses challenges in objectively comparing NER approaches and analyzing their errors.

To address this issue, we present a thorough re-annotation effort that corrects **6.4%** of the label mistakes in the ANERcorp dataset and produces a cleaner version of the dataset named (CLEANANERCorp) that significantly improves annotation quality and consistency.

To the best of our knowledge, this is the first study that systematically handles label mistakes in the ANERcorp dataset. We conducted extensive experiments on both the original ANERcorp dataset and our corrected dataset CLEANANERCorp and achieved superior results.

The contributions of this study are as follows:

- We present CLEANANERCorp, a clean version of ANERcorp that includes corrected, consistent and reliable NER annotations in both splits, where (6.45%) of the training set and (6.16%) of the test set of the ANERcorp have been updated.
- We re-evaluated the popular Arabic NER models with CLEANANERCorp and achieved a marginally high increase with the F1 score results, which is about (7.23%).
- We re-evaluated the popular Cross-lingual NER models that achieved state-of-the-art performance with the corrected test set and achieved higher results.

CLEANANERCorp is publicly available to encourage the community to use it and to improve its quality further[1].

## 2. ANERcorp Overview

ANERcorp is one of the earliest and most widely adopted NER corpora for Arabic. It was published in 2007 and has since become the standard in the Arabic NER literature. ANERcorp comprises two corpora for training and one for testing. The total number of articles included 316 from different newspapers.

The dataset annotation guidelines followed in the ANERcorp dataset were based on MUC Conventions (Sang & De Meulder, 2003). Following this guideline, the dataset was tagged with four entities: *person (PER), location (LOC), organization (ORG), and miscellaneous (MISC)*. The tagging scheme is the inside–outside–beginning (IOB) scheme originally proposed by ((Ramshaw & Marcus, 1999). Therefore, any word on the text should be annotated as one of the following tags:

- B-PER: The Beginning of the name of a person.[2]
- I-PER: The continuation (Inside) of the name of a person.
- B-LOC: The Beginning of the name of a location.

---

[1] Github link: https://github.com/iwan-rg/CLEANANERCorp
[2] The original dataset used B-PERS instead of B-PER and I-PERS instead of I-PER in the annotation. We re-annotate the dataset with the same original tags in the dataset but refer to them as B-PER and I-PER in this paper.

- I-LOC: The Inside of the name of a location.
- B-ORG: The Beginning of the name of an organization.
- I-ORG: The Inside of the name of an organization.
- B-MISC: The Beginning of the name of an entity that does not belong to any of the previous classes (miscellaneous).
- I-MISC: The Inside of the name of an entity that does not belong to any of the previous classes.
- O: The word is not a named entity (Other).

The dataset contains (150,286) tokens and (32,114) types, which makes the ratio of tokens to types is (4.67). The distributions of the different tags are listed in Table 1.

| Class | Ratio |
|---|---|
| PER | 39% |
| LOC | 30.4% |
| ORG | 20.6% |
| MISC | 10% |

Table 1 Ratio of phrases by classes

In 2020, the CAMeL Lab (Obeid et al., 2020) released a new version of ANERcorp, where they split the data and performed minor corrections agreed upon with the original author.

The changes from the original dataset include the following:

- Correct minor tag spelling errors.
- Convert the middle periods (·) and bold periods (•) to regular periods (.).
- Remove the blank Unicode character (\u200F).
- Add sentence boundaries after sequences of one or more periods.
- Split the dataset sequentially. The sentences containing the first 5/6 of the words go to training, and the rest go to testing. The training split had 125,102 words, and the test split had 25,008 words.

However, no previous efforts have been made to correct tagging errors and mislabeling in the ANERcorp dataset. We have carefully reviewed the original ANERcorp and identified the different types of labeling errors. They are listed below with examples:

### A. Label Inconsistency

Some tokens were tagged differently for each sentence. For example, (الدولارات، جنيه استرليني) has been tagged sometimes as MISC and sometimes as O. Also, (الضفة الغربية) has been tagged as LOC and O in different sentences.

### B. Wrong Labels

In Figure 1, the word "المتحدة" has been tagged as B-ORG while it should be tagged as I-LOC.

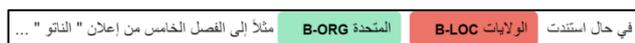

Figure 1 An Example of a Wrong Label

### C. MISC tag Ambiguity

As the dataset follows the same classes that were defined in the MUC-6 Conventions (Sang & De Meulder, 2003) (Organization, Location, Person, and Miscellaneous), the MISC tag was not covered correctly and many MISC entities were tagged as O.

### D. Sentence Beginning Ambiguity

We noticed an ambiguity in the first words of many sentences where the correct label was not clear. Figure 2 shows an example of such a sentence where the word (برند) has been tagged as (B-PER) and the meaning of the word is not clear.

### E. Typographical Errors

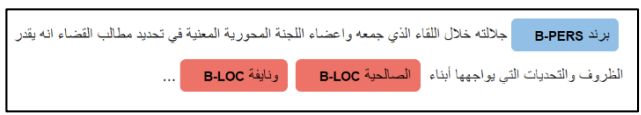

Figure 2 Sentence Beginning Ambiguity Example

In addition to tagging errors, we noticed some typographical errors in the dataset. The dataset was written in two columns, where each word was placed on a separate line with its tag. We encountered two words attached to each other in one line without space. For example: (فيهاالبلدان، التفسيرالنصى، ولمافشلت، وراءوالدهم، المصادرالتاريخية، إنسبعةعراقيين، أكبرمحافظة).

## 3. Reannotation Process

The reannotation process was conducted in four distinct phases.

### 3.1 Annotation Guideline Definition

The ANERcorp annotation guidelines are based on MUC-6 Conventions (Sang & De Meulder, 2003). Following this guideline, we defined four entities: person (PER), location (LOC), and organization (ORG), using an extra miscellaneous (MISC) type to deal with entities that do not fall into these categories. The guidelines were refined to suit the Arabic language. For example, we consider prefixes to be part of the entity names. For example: (منظمة بورصة نيويورك), (شركة النفط النيجيرية), (الامم المتحدة).

We developed a special handling for ambiguities in the guidelines to resolve cases that were not clear during the revision. In most cases, we assigned a tag that matched the context of the sentence. Following (Rücker & Akbik, 2023), we decided to tag the national sport team with ORG instead of LOC (المنتخب المصري، المنتخب السعودي). Political houses were also tagged as LOC (البيت الأبيض، الكرملين). We have noticed inconsistency in tagging the currency, sometimes as MISC and sometimes as O or LOC. Following CoNLL tagging, we decided to label the currency and physical units as O instead of MISC.

## 3.2 Automatic Error Detection with CLEANLAB

CLEANLAB[3] is a framework that automatically detects label issues in a machine learning dataset using confident learning (Wang & Mueller, 2022). This framework uses existing models to detect dataset problems that can be fixed to train even better models. We utilized CLEANLAB as a first round to check the number of issues in the dataset. We detected (1945) issues. These issues have been manually investigated and corrected.

## 3.3 Manual Re-annotation

An annotator was hired to manually re-annotate all the entities in the dataset. The annotator was provided with guidelines and encouraged to use search engines and Wikipedia for suspicious token spans. The dataset was split into nine files for ease of handling.

## 3.4 Final Revision

After re-annotating all the tokens, a final round of revision has been conducted by the annotator and the author to resolve any ambiguity and inconsistency in the updated tags.

Finally, we corrected and added a total of **9605** label mistakes, which is approximately **6.4%** of the dataset.

## 4. Evaluation

### 4.1 Dataset Statistics

All labeling errors and typographical errors detected were resolved. The following subsections present some statistics on the data.

### A. Label Distribution

Tables 2 and 3 compare the total count of annotated named entities and the distribution across the four classes for CLEANANERCorp and the original ANERcorp. We observe that CLEANANERCorp has a slightly higher number of ORG and MISC entities than the base version. This stems from a more consistent use of ORG labels for sports teams referred to by their geographic name, and a more consistent use of MISC for adjectives and entity types, such as sports leagues and events.

|       | ANERcorp |       | CLEANANERCorp |      |
|-------|----------|-------|---------------|------|
| Class | #        | %     | #             | %    |
| PER   | 1499     | 5.99  | 1508          | 6.03 |
| LOC   | 751      | 3.00  | 802           | 3.21 |
| ORG   | 725      | 2.90  | 1035          | 4.14 |
| MISC  | 400      | 1.60  | 1090          | 4.36 |
| O     | 21633    | 86.50 | 20573         | 82.27|
| Total | 25008    |       | 25008         |      |

Table 2 Statistics of test set entities in ANERcorp vs. CLEANANERCorp datasets.

[3] https://github.com/cleanlab

|       | ANERcorp |        | CLEANANERCorp |        |
|-------|----------|--------|---------------|--------|
| Class | #        | %      | #             | %      |
| PER   | 4926     | 3.94%  | 4906          | 3.92%  |
| LOC   | 4301     | 3.44%  | 4610          | 3.68%  |
| ORG   | 2691     | 2.15%  | 4394          | 3.51%  |
| MISC  | 1263     | 1.01%  | 5263          | 4.21%  |
| O     | 111921   | 89.46% | 105929        | 84.67% |
| Total | 125102   | 100%   | 125102        | 100%   |

Table 3 Statistics of entities of the training set in ANERcorp vs. CLEANANERCorp datasets.

### B. Labels Changed

Table 4 shows the extent of the label updates introduced compared to the original dataset. A total of (**9605**) labels were modified from the original dataset, which is (**6.4%**) of the total dataset. Tables 5 and 6 further examine the update details for each data split.

|           | CLEANANERCorp |       |
|-----------|---------------|-------|
|           | #             | %     |
| Changed   | 9605          | **6.4%** |
| Unchanged | 140505        | 93.6% |
| Total     | 150110        | 100%  |

Table 4 NER labels updated in CLEANANERCorp datasets.

|                  | CLEANANERCorp train set |        |
|------------------|-------------------------|--------|
|                  | #                       | %      |
| Label Corrected  | 1667                    | 1.33%  |
| Label Added      | 6397                    | 5.11%  |
| Label Unchanged  | 117038                  | 93.55% |
| #Entities        | 125102                  | 100%   |

Table 5 NER labels in the CLEANANERCorp train set according to the type of change.

|                  | CLEANANERCorp test set |        |
|------------------|------------------------|--------|
|                  | #                      | %      |
| Label Corrected  | 369                    | 1.48%  |
| Label Added      | 1172                   | 4.69%  |
| Label Unchanged  | 23467                  | 93.84% |
| #Entities        | 25008                  | 100%   |

Table 6 NER labels in the CLEANANERCorp test set according to the type of change.

## 5. Experiments

To determine the extent to which our relabeling effort affects model performance, we re-evaluated a set of NER models on CLEANANERCorp and ANERcorp in two different settings: monolingual and cross-lingual transfer.

Currently, fine-tuning large pre-trained language models has achieved state-of-the-art performance on both monolinguals (Antoun et al., 2021a, 2021b) and cross-lingual NER (Hu et al., 2020; Lan et al.,

2020). Therefore, we selected pre-trained language models from the literature that report state-of-the-art results on Arabic and English-Arabic cross-lingual transfer and re-evaluated them on different dataset versions for the NER task.

For the cross-lingual transfer, we experimented with a zero-shot cross-lingual transfer from English to Arabic, where the model was trained on English data and tested on Arabic. We used the CoNLL2003 dataset (Sang & De Meulder, 2003) for training and validation.

Although there are other published results (Abdul-Mageed et al., 2021; Khalifa & Shaalan, 2019) with higher SOTA, they reported the results on different data splits and tested the models without the MISC tag, focusing only on three tags: person (PER), location (LOC), and organization (ORG), while setting other labels to the unnamed entity (O).

### 5.1 Reference Models

We re-evaluated state-of-the-art Arabic and multilingual language models on the CLEANANERCorp and ANERcorp datasets.

For the Arabic pretrained language models, we re-evaluated the following:

- **ARABERT**v0.2 base (Antoun et al., 2021a): The state-of-the-art Arabic-specific BERT model for various Arabic IE tasks. The model contained 24 layers of encoders stacked on top of each other, 16 self-attention heads, and a hidden size of 1024.
- **ARBERT** (Abdul-Mageed et al., 2021): Arabic-specific Transformer LMs pre-trained on very large and diverse datasets, including MSA as well as Arabic dialects.
- **AraELECTRA** (Antoun et al., 2021b): A pretrained ELECTRA model on a large-scale Arabic dataset.

For the cross-lingual experiments, we re-evaluated

- **mBERT** (Devlin et al., 2019): Multilingual BERT pretrained on Wikipedia of 104 languages using masked language modelling (MLM).
- **XLM-RoBERT** (XLM-R) (Conneau et al., 2020): A transformer-based multilingual masked language model pre-trained on text in 100 languages that obtains state-of-the-art performance on different cross-lingual tasks.
- **GigaBERT** (Lan et al., 2020): A bilingual BERT for English-to-Arabic cross-lingual transfer trained on newswire English and Arabic text from the Gigaword dataset in addition to Wikipedia and Web crawl data.

**Hyperparameter:** For monolingual fine-tuning experiments, we followed the same hyperparameter reported by ((Antoun et al., 2021b) where all the models were fine-tuned with batch size set to (32), maximum sequence length of (256), and learning rates (5e-5). For cross-lingual fine-tuning, we followed the same hyperparameters reported by ((Hu et al., 2020), where mBERT was fine-tuned for two epochs, with a training batch size of (32) and a learning rate of (2e-5), and XLM-R was fine-tuned for two epochs with a learning rate of 3e-5 and size of 16. All hyperparameter tuning for the cross-lingual experiment was performed on the English validation data.

### 5.2 Monolingual Results

The experimental results of the tested models for the different dataset versions are listed in Table 7. F1-score was averaged over three runs with different seeds for each experimental setting.

| Model | Train/Test : **ANERcorp** | Train/Test : **CLEANANERCorp** |
|---|---|---|
| AraBERT v2 | 0.83 | **0.89** |
| ARBERT | 0.83 | **0.89** |
| AraELECTRA | 0.82 | **0.87** |

Table 7 Average F1 score of fine-tuning Arabic LMs on ANERcorp vs. CLEANANERCorp datasets.

The results show that CLEANANERCorp achieved marginally higher performance on all tested models compared to the original dataset, which indicates that our relabeling effort successfully improved label quality and consistency.

AraBERT F1 score has increased by (7.23%) from (0.83) to (0.89) after re-annotation. Table 8 shows a detailed comparison of each entity type in terms of Precision, Recall and F1-score for the AraBERT model on the two versions of the datasets.

We can see that all the F1 scores increased after correction, and the highest gain in entity F1 score was from the MISC and ORG labels, where the F1 score increased by (26.47%) and (16%), respectively.

|  | **ANERcorp** | | | **CLEANANERCorp** | | |
|---|---|---|---|---|---|---|
|  | Prec | Rec | F1 | Prec | Rec | F1 |
| LOC | 0.89 | 0.93 | 0.91 | 0.94 | 0.92 | 0.93 |
| MISC | 0.73 | 0.63 | 0.68 | 0.85 | 0.86 | 0.86 |
| ORG | 0.76 | 0.73 | 0.75 | 0.85 | 0.87 | 0.86 |
| PER | 0.88 | 0.84 | 0.86 | 0.93 | 0.90 | 0.92 |
| Overall | 0.84 | 0.82 | 0.83 | 0.89 | 0.89 | 0.89 |

Table 8 Entity-based precision, recall, and F1 score of fine-tuned AraBERT on ANERcorp vs. CLEANANERCorp datasets.

### 5.3 Cross-Lingual Zero-Shot Transfer Results

Table 9 reports the average F1 scores over three runs with different seeds for each experimental setting.

From the results in Table 9, we can observe a high increase in F1 scores when transferring to the corrected dataset compared to those on the original test set.

| Model | Train: Conll2003 Test: ANERcorp | Train: Conll2003 Test: CLEANANERCorp |
|---|---|---|
| mBERT-base | 0.46 | **0.48** |
| XLM-R-base | 0.52 | **0.62** |
| XLM-R-Large | 0.53 | **0.62** |
| GigaBERT | 0.61 | **0.72** |

Table 9 Average F1 Scroe of Cross-lingual transfer on the ANERcorp vs. CLEANANERCorp datasets.

For example, fine-tuning XLM-r-base achieved (19.23%) increase from the (0.52) to (0.62) F1-score. Table 10 shows the F1 score per entity type, where we can see a high increase in the MISC label F1 score from (0.08) to (0.57), which justifies the increase in the overall score.

|  | ANERcorp | | | CLEANANERCorp | | |
|---|---|---|---|---|---|---|
|  | Prec | Rec | F1 | Prec | Rec | F1 |
| LOC | 0.63 | 0.72 | 0.68 | 0.61 | 0.70 | 0.65 |
| MISC | 0.05 | 0.19 | 0.08 | 0.59 | 0.56 | 0.57 |
| ORG | 0.41 | 0.54 | 0.46 | 0.44 | 0.53 | 0.48 |
| PERS | 0.61 | 0.71 | 0.66 | 0.70 | 0.70 | 0.70 |
| Overall | 0.43 | 0.63 | 0.51 | 0.59 | 0.63 | 0.61 |

Table 10 Entity-based precision, recall, and F1 score of the Cross-lingual Transfer of XLM-R on ANERcorp vs. CLEANANERCorp dataset.

The above results indicate that CLEANANERCorp is more consistent with the CONLL2003 dataset and can be used to reflect the accuracy of the Cross-lingual Zero-Shot models more stably.

## 6. Conclusion

We presented CLEANANERCorp, a corrected and cleaner version of the widely adopted Arabic NER benchmark dataset ANERcorp. Our re-annotation updated (6.4%) the labels in the original dataset. Our evaluation of monolingual and cross-lingual NER language models achieved higher performance and strongly indicated that the overall annotation quality and consistency were significantly improved. Therefore, we contribute to improving the quality of the public Arabic NER datasets with updated and more consistent NER labels.

## 7. Acknowledgments

We would like to Thank Dr. Mustafa Jarrar, Sana Ghanem and Tymaa Hammouda from SinaLab for Computational Linguistics and Artificial Intelligence[4] to their contribution in explaining Wojood Model guidelines and annotating the ANERCorp with Wojood model (Jarrar et al., 2022).

---
[4] https://sina.birzeit.edu/index.html